\documentclass[journal]{IEEEtran}

\usepackage{array,amsthm}
\usepackage[dvipsnames]{xcolor}   
\usepackage{hyperref}             
\usepackage{graphicx}

\usepackage{tikz}
\usetikzlibrary{arrows.meta, positioning, fit, backgrounds, calc}
\usepackage{amsmath,amssymb,amsfonts}
\usepackage{mathtools}
\usepackage{algorithm}
\usepackage{algorithmic}
\usepackage{graphicx}
\usepackage{textcomp}
\usepackage{xcolor}
\usepackage{booktabs}
\usepackage{siunitx}
\usepackage{multirow}
\usepackage{url}
\usepackage{xcolor}
\usepackage{cite}

\definecolor{cIn}  {HTML}{1B4F72}   
\definecolor{cEmb} {HTML}{6C3483}   
\definecolor{cPE}  {HTML}{0E6655}   
\definecolor{cEnc} {HTML}{7E5109}   
\definecolor{cPool}{HTML}{212F3D}   
\definecolor{cMu}  {HTML}{922B21}   
\definecolor{cSig} {HTML}{196F3D}   
\definecolor{cPi}  {HTML}{1A5276}   
\definecolor{cOut} {HTML}{0B5345}   
\definecolor{cGray}{HTML}{717D7E}   
\definecolor{cDark}{HTML}{2C3E50}   

\tikzset{
  pipebox/.style 2 args={
    draw=#1!80!black, fill=#1!10!white, line width=1.2pt,
    rounded corners=4pt, minimum width=#2, inner sep=6pt,
    align=center, font=\small\sffamily\bfseries, text=black
  },
  mdnbox/.style 2 args={
    draw=#1!80!black, fill=#1!10!white, line width=1.2pt,
    rounded corners=3.5pt, minimum width=#2, minimum height=1.1cm,
    inner sep=5pt, align=center,
    font=\footnotesize\sffamily\bfseries, text=black
  },
  dbox/.style={
    draw=cEnc!80!black, fill=cEnc!8!white, line width=1.2pt,
    rounded corners=3pt, inner sep=4pt,
    align=center, font=\scriptsize\sffamily\bfseries, text=black
  },
  circ/.style={
    draw=cDark!60, circle, fill=white, minimum size=0.4cm,
    inner sep=0pt, font=\small\sffamily\bfseries,
    text=black, line width=0.7pt
  },
  far/.style={-Stealth, line width=1.0pt,  color=black,      shorten >=2pt, shorten <=2pt},
  bar/.style={-Stealth, line width=0.85pt, color=black,   shorten >=2pt, shorten <=2pt},
  dar/.style={-Stealth, line width=0.65pt, color=cEnc!100,    shorten >=1pt, shorten <=1pt},
  res/.style={-Stealth, line width=0.6pt,  color=cGray!100,   shorten >=1pt, shorten <=1pt},
  shp/.style ={font=\scriptsize\sffamily, color=Black,    align=center},
  stag/.style={font=\tiny\sffamily\bfseries, color=black, align=center}
}

\usepackage{cite}
\usepackage{amsmath,amssymb,amsfonts}
\usepackage{algorithmic}
\usepackage{graphicx}
\usepackage{textcomp}
\usepackage{xcolor}
\def\BibTeX{{\rm B\kern-.05em{\sc i\kern-.025em b}\kern-.08em
    T\kern-.1667em\lower.7ex\hbox{E}\kern-.125emX}}

\newcommand{\bv}{\mathbf{v}}

\newcommand{\bd}{\boldsymbol{\delta}}

\newcommand{\Tobs}{T_{\mathrm{obs}}}
\newcommand{\Tpred}{T_{\mathrm{pred}}}
\newcommand{\dmodel}{d_{\mathrm{model}}}

\begin{document}

\title{AeroCast: Probabilistic 3D Trajectory Prediction for
Non-Cooperative Aerial Obstacles via
Transformer-MDN Architecture}

\author{Syed~Izzat~Ullah~and~Jos\'e~Baca%
\thanks{The authors are with the Department of Engineering, Texas A\&M University--Corpus Christi, Corpus Christi, TX 78412 USA (e-mail: sizzatullah@islander.tamucc.edu; jose.baca@tamucc.edu).}%
\thanks{Corresponding author: Jos\'e Baca.}}

\maketitle

\begin{abstract}Autonomous aerial vehicles operating in shared airspace must predict the
future positions of non-cooperative obstacles to plan evasive maneuvers
before a collision becomes unavoidable.  Unlike cooperative systems that
share intent, non-cooperative obstacles such as birds, uncontrolled
drones, or debris exhibit multi-modal motion that deterministic
predictors cannot adequately represent.  Existing methods either rely on
recurrent encoders that propagate temporal information sequentially,
limiting their ability to capture long-range kinematic precursors of
maneuver initiation, or produce point forecasts that provide no
distributional information to downstream planners. This paper presents
AeroCast, a probabilistic trajectory prediction framework that combines a Transformer encoder with a
Mixture Density Network output head to predict per-timestep Gaussian
mixture distributions over future three-dimensional displacements. A
translation-invariant consecutive displacement encoding and a
calibration-oriented training objective address the input design and
mode-degeneracy challenges specific to mixture-based aerial trajectory
prediction. On a hybrid real-and-synthetic quadrotor corpus spanning
nine motion categories, AeroCast reduces Average Displacement Error and
Final Displacement Error by approximately 50\% relative to the
baselines over a five-second horizon, and achieves the lowest negative
log-likelihood and Continuous Ranked Probability Score among all
compared methods. Ablation analysis
identifies velocity input and model capacity as the primary contributors
to prediction quality, and positional encoding as essential for
long-horizon trajectory coherence.  AeroCast inference completes in 0.1\,ms per
sample, compatible with real-time onboard deployment at 100\,Hz.
\end{abstract}


\section{INTRODUCTION}
T{\scshape he} proliferation of unmanned aerial vehicles (UAVs) in
shared airspace introduces a core safety requirement: an autonomous
vehicle must anticipate the future positions of surrounding
non-cooperative obstacles to plan evasive maneuvers before a collision
becomes geometrically unavoidable.  Unlike cooperative multi-agent
systems that share intent through communication protocols,
non-cooperative obstacles such as birds, uncontrolled drones, or
wind-borne debris provide no advance notice of their behavior.  Reactive
strategies that respond only to instantaneous obstacle positions offer
insufficient safety margins at typical flight speeds, where the interval
between detection and potential collision may be shorter than the
vehicle's minimum stopping distance~\cite{ego_swarm}.  Proactive
avoidance requires a predictor that maps a finite observation history to
a distribution over future positions. This task is more challenging in
the aerial domain than in ground-domain forecasting~\cite{chandran2023collisionavoidance}. Aerial obstacles
move in unconstrained three-dimensional space without the road geometry
or pedestrian social norms that regularize planar motion, and their
behavior is inherently multi-modal: an approaching obstacle may continue
straight, bank sharply, or climb, each with non-trivial probability.
 
Existing methods address these challenges only partially.  Physics-based
estimators such as Kalman filters and Interacting Multiple Model (IMM)
trackers assume predefined motion models (constant velocity, coordinated
turn) that poorly capture the erratic behavior of non-cooperative
obstacles~\cite{syed2025pofmader, cmas}.  Learning-based recurrent
architectures, notably the GRU-based VECTOR model for 3-D UAV
prediction~\cite{nacar2025vector}, achieve fast inference suitable for
embedded deployment but produce deterministic point forecasts.  A
single-trajectory output gives a downstream planner no information about
whether the prediction is confident or whether the obstacle is at a behavioral branch point, both situations are indistinguishable in the
output.  Transformer-based encoders offer a structural alternative: the
self-attention mechanism provides every observation token with direct
access to every other token in the window, shortening the information
path between temporally distant timesteps to $O(1)$ and removing the
sequential bottleneck inherent in recurrent processing~\cite{vaswani2017attention}.  To date, no
prior work has applied a Transformer encoder specifically to
non-cooperative aerial obstacle prediction with a multi-modal
probabilistic output head targeting uncertainty.
 
This paper presents AeroCast, a probabilistic trajectory prediction
framework that pairs a Pre-LN Transformer encoder with a Mixture
Density Network (MDN) output head to predict per-timestep Gaussian
mixture distributions over future three-dimensional displacements.  The
input representation encodes consecutive positional displacements
normalized by a scale that enforces translation invariance and produces bounded inputs suited to Gaussian
mixture modeling.  The training objective combines negative
log-likelihood with a mode-anchoring mean-squared-error term and a
variance floor matched to the sensor noise level, which prevents the
mode degeneracy that arises under pure NLL optimization with multiple
mixture components.  AeroCast is evaluated on a hybrid real-and-synthetic
quadrotor trajectory corpus spanning nine kinematically diverse motion
categories, with quantitative comparisons against four baselines at
matched parameter budgets, per-category accuracy breakdowns, and a
systematic ablation of all major design choices.

A supplementary video demonstrating AeroCast predictions
across various trajectories is available at
\url{https://syediu.github.io/assets/img/aerocast_circle.mp4}. 

The main contributions are:
\begin{enumerate}
  \item \textbf{A probabilistic Transformer-MDN predictor for
    non-cooperative aerial obstacles.}  AeroCast combines a Pre-LN
    Transformer encoder with MDN output heads to produce per-timestep
    Gaussian mixtures over future 3-D displacements, reducing ADE and
    FDE by approximately 50\% relative to the strongest recurrent
    baseline and achieving the lowest NLL and CRPS among all compared
    methods.
 
  \item \textbf{A translation-invariant displacement encoding and
    calibration-oriented training protocol.}  The consecutive
    displacement representation provides translation invariance, bounded
    input distributions, and Gaussian compatibility.  The combined
    NLL and mode-anchoring objective with a physically grounded sigma
    floor prevents variance collapse and produces uncertainty estimates
    through Expected Calibration Error.
 
      \item \textbf{A comprehensive experimental evaluation and benchmarking.} We conduct quantitative comparisons against four trained baselines at matched parameter budgets, a per-trajectory-type accuracy breakdown across nine kinematically diverse motion categories, a prediction horizon analysis tracking error growth from 1 to 5 seconds, and a systematic ablation of all major architectural and training choices.
  \item \textbf{A hybrid real-and-synthetic benchmark for non-cooperative aerial obstacle prediction.} We contribute a dataset of 90,116 sliding-window sequences from 113 quadrotor flight recordings across nine kinematically diverse motion categories that combines sub-millimeter Vicon ground truth with synthetic augmentation derived from a parametric trajectory generator. This dataset, together with a standardized evaluation protocol covering ADE, FDE, best-of-K, NLL, and ECE metrics, to our knowledge, constitutes the first structured benchmark for 3-D non-cooperative aerial obstacle trajectory prediction.
\end{enumerate}
\begin{figure}[t]
    \centering
    \includegraphics[width=\columnwidth]{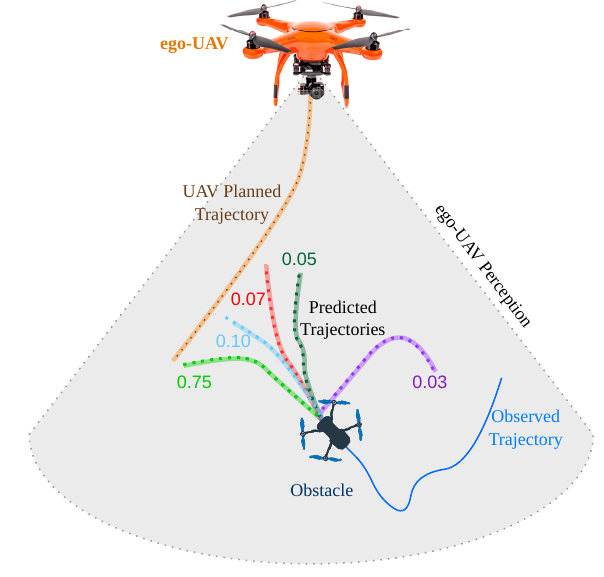}
    \caption{A typical scenario where an ego-UAV encounters a non-cooperative dynamic obstacle (an uncontrolled drone). The ego-UAV predicts the K-future most plausible predictions of the obstacle}
    \label{fig:fig1}
\end{figure}

\section{RELATED WORK}
\label{sec:related}
\subsection{Classical Estimation and Recurrent Forecasting for Aerial Trajectories}
The classical approach to aerial obstacle tracking relies on Bayesian state estimators that maintain a finite bank of kinematic hypotheses updated recursively as observations arrive. Kalman filters and their multi-model extensions, including Interacting Multiple Model trackers, are computationally efficient and interpretable, but their predictive quality is bounded by the fidelity of the assumed motion models to the underlying dynamics \cite{syed2025pofmader}. For non-cooperative aerial obstacles exhibiting abrupt heading reversals, irregular vertical excursions, or speed-coupled turn maneuvers, no compact set of constant-velocity or coordinated-turn hypotheses adequately spans the behavioral envelope, and the estimator systematically lags behind the true state evolution rather than anticipating it~\cite{syed2023msWork, dqnNeimat2021}.

Learning-based recurrent architectures displaced hand-crafted motion models by
extracting temporal structure directly from data. LSTM, GRU, and
bidirectional GRU variants has been used as a standard components in trajectory
forecasting pipelines. For aerial trajectory prediction specifically, VECTOR~\cite{nacar2025vector} introduced a velocity-enhanced GRU formulation for 3D UAV prediction, that achieves fast inference suitable for onboard deployment. However, the sequential update mechanism imposes an information path length proportional to observation length, so early kinematic signatures of an impending maneuver must
propagate through every intermediate hidden transition before influencing the forecast. Moreover, all recurrent methods cited above produce deterministic point forecasts, providing no information to a downstream planner about whether the prediction reflects high confidence or an imminent behavioral bifurcation.

\subsection{Transformer-Based Trajectory Forecasting}

The Transformer formulation of Vaswani et al.~\cite{vaswani2017attention}
replaced recurrence with self-attention, allowing every observation token
to interact directly with every other within the window. Global
connectivity shortens the information path length to a constant number of
steps between any two timesteps. Stability
of deep encoder stacks is improved by pre-layer normalization, which
places normalization inside the residual branch and reduces the
optimization fragility of post-normalization
Transformers~\cite{xiong2020layer}. These properties have been exploited
in vehicle and pedestrian forecasting through hierarchical and vectorized
attention models~\cite{yuan2021agentformer,zhou2022hivt}. To our knowledge, no prior work applies a Transformer encoder
specifically to non-cooperative aerial obstacle prediction, and none couples
such an encoder with a multi-modal probabilistic output head targeting
calibrated uncertainty estimation. This work adopts
the same architectural motivation but applies it to non-cooperative
aerial obstacles, using a Transformer encoder over a 2~s history of 3-D
position and velocity so the model can attend globally to maneuver
precursors over the full observation horizon.

\subsection{Probabilistic Prediction and Calibration}

Deterministic forecasts collapse distinct future behaviors into a single
mean path, which is insufficient for risk-aware planning that treats
uncertainty as a decision variable.  Multi-modal output representations
have become standard in ground-domain forecasting:
Social-GAN~\cite{gupta2018socialgan} promotes diverse futures through
adversarial training and a variety loss;
Trajectron++~\cite{salzmann2020trajectron} combines a graph-structured
recurrent backbone with conditional latent variables; and
AgentFormer~\cite{yuan2021agentformer} extends multi-modal prediction
to Transformer architectures with social context.  These methods
establish that single-hypothesis outputs are inadequate in dense or
interactive scenes, but they were designed around ground-domain
constraints (map topology, pedestrian norms) that have no counterpart in
free-flight aerial prediction.

A separate question is whether the predicted uncertainty is
\emph{calibrated}, which means, whether a model's stated confidence
intervals match empirical coverage rates.  Guo et
al.~\cite{guo2017calibration} showed that modern deep networks are
systematically miscalibrated and formalized Expected Calibration Error
(ECE) as a summary statistic for the confidence--coverage mismatch.
Lakshminarayanan et al.~\cite{lakshminarayanan2017ensembles}
demonstrated that deep ensembles produce well-calibrated uncertainty
through multi-model averaging, at the cost of multiple forward passes
that increase computational and memory requirements on embedded
platforms.  Kendall and Gal~\cite{kendall2017uncertainty} distinguished
aleatoric uncertainty (irreducible measurement noise and behavioral
variability) from epistemic uncertainty (reducible with more data or
richer models), a decomposition that clarifies which source of
uncertainty a given architecture can represent.  On the training side,
Gneiting and Raftery~\cite{gneiting2007proper} proved that strictly
proper scoring rules, including negative log-likelihood, are the correct
objective when the output is a probability distribution; mean squared
error is not proper at the distribution level and can drive variance
collapse.  Ivanovic et al.~\cite{ivanovic2022propagating} extended this
perspective to trajectory forecasting by analyzing how predictive
uncertainty propagates along the forecast horizon.

Most multi-modal trajectory predictors evaluate probabilistic quality
only indirectly, through best-of-$K$ error or diversity measures, and
none of the cited methods enforces calibration during training.
AeroCast encodes aleatoric ambiguity in a per-timestep Gaussian mixture
trained with NLL as the primary objective, augmented by a mode-anchoring
MSE term and a sigma floor matched to the sensor noise level.  This
design targets single-pass inference compatible with onboard deployment
and produces uncertainty estimates that can be assessed through ECE.

\subsection{Datasets for Non-Cooperative Aerial Prediction}

Public trajectory datasets at the scale and structure available for
ground-domain prediction do not exist for non-cooperative 3-D aerial
obstacles.  Pedestrian benchmarks such as
ETH~\cite{pellegrini2009you} and driving corpora such as
nuScenes~\cite{caesar2020nuscenes} and the Waymo Open Motion
Dataset~\cite{ettinger2021large} contain planar, map-constrained, or
cooperative traffic structure absent from free-flight aerial settings.
Existing aerial corpora focus on cooperative multi-UAV operations or
structured air-traffic recording, underrepresenting the abrupt and
vertically coupled maneuvers relevant to collision avoidance in shared
airspace.  SynTraG~\cite{syed2026syntrag} introduced a parametric
generator for non-cooperative dynamic obstacles in UAV navigation.  The
present work combines a Vicon-collected real-flight corpus~\cite{ullah2026nanobench} with
synthetic~\cite{syed2026syntrag} augmentation derived from an extension of these primitives,
coupling real 3-D obstacle motion with synthetic samples that share the
same physical and noise assumptions used by the predictor~\cite{autonomousSR2023, 2stageDL2026, realtime_detection2026}.

Table~\ref{tab:related} summarizes the positioning of AeroCast relative
to representative prior work across the design dimensions identified
above.

\begin{table}[t]
  \centering
  \caption{%
    Positioning of AeroCast relative to representative trajectory prediction
    methods. \checkmark: property is present; {---}: absent or not reported.
    \emph{Attn.}: Transformer self-attention encoder;
    \emph{Multi-modal}: explicit multi-hypothesis output distribution;
    \emph{Calib.}: calibration enforced during training or reported via ECE;
    \emph{Aerial}: targets non-cooperative 3-D aerial obstacles.%
  }
  \label{tab:related}
  \setlength{\tabcolsep}{4.5pt}
  \begin{tabular}{@{}lccccc@{}}
    \toprule
    {Method} & {3-D} & {Attn.} & {Multi-modal} & {Calib.} & {Aerial} \\
    \midrule
    Social-GAN~\cite{gupta2018socialgan}              & ---        & ---        & \checkmark & ---        & ---        \\
    Trajectron++~\cite{salzmann2020trajectron}     & ---        & ---        & \checkmark & ---        & ---        \\
    AgentFormer~\cite{yuan2021agentformer}         & ---        & \checkmark & \checkmark & ---        & ---        \\
    VECTOR~\cite{nacar2025vector}                  & \checkmark & ---        & ---        & ---        & \checkmark \\
    \textbf{AeroCast (ours)}                       & \checkmark & \checkmark & \checkmark & \checkmark & \checkmark \\
    \bottomrule
  \end{tabular}
\end{table}

\section{PROBLEM FORMULATION}

The trajectory prediction problem is formulated from the perspective of
an ego aerial vehicle observing external dynamic obstacles in its
vicinity.  Let
$\mathcal{X} = \{(\mathbf{x}_t, \mathbf{v}_t)\}_{t=1}^{\Tobs}$ denote
the observed trajectory history of an obstacle, where
$\mathbf{x}_t \in \mathbb{R}^3$ is the three-dimensional position
relative to the ego vehicle at time~$t$ and
$\mathbf{v}_t \in \mathbb{R}^3$ is the corresponding velocity vector.
Positions $\mathbf{x}_t = [x_t, y_t, z_t]^\top$ are expressed in a
world-fixed inertial frame with the $z$-axis opposing gravity; for the
experimental validation this frame coincides with the Vicon system
origin.  Velocities are computed through centered finite differences with
Gaussian smoothing:
\begin{equation}
  \mathbf{v}_t
    = \frac{\mathbf{x}_{t+1} - \mathbf{x}_{t-1}}{2\Delta t}
      \ast \mathcal{G}(\sigma)
  \label{eq:velocity}
\end{equation}
where $\Delta t$ is the sampling interval and
$\mathcal{G}(\sigma)$ is a Gaussian kernel with
$\sigma = 0.5\,\Delta t$, chosen to balance noise suppression and
temporal resolution.

The objective is to predict the obstacle's future trajectory
$\mathcal{Y} = \{\mathbf{y}_t\}_{t=\Tobs+1}^{\Tobs+\Tpred}$, where
$\mathbf{y}_t \in \mathbb{R}^3$ is the predicted position at time~$t$.
The prediction is model-free; no knowledge of the obstacle's dynamics,
control law, or physical parameters is assumed.  The obstacle may be a
bird, an autonomous drone, a remotely piloted vehicle, or any other
aerial agent.  This model-agnostic requirement motivates a
learning-based approach that infers motion patterns purely from observed
kinematics.

\subsection{Temporal Specifications}

Observations span $\Tobs = 2.0$\,s sampled at 10\,Hz, yielding input
sequences of length~20.  Predictions extend $\Tpred = 5.0$\,s into the
future (50~timesteps at 10\,Hz).  The 10\,Hz rate matches typical
tracking-system update frequencies and provides sufficient temporal
resolution to capture the motion dynamics of small agile platforms.  The
5\,s horizon provides a collision avoidance planner with enough lead time
to identify conflicts and execute evasive maneuvers.

\subsection{Consecutive Displacement Encoding}
\label{subsec:disp_enc}

Raw position coordinates $\mathbf{x}_t$ are translation-variant where the
same motion pattern yields different input values depending on where in
the workspace the obstacle is located.  To remove this redundancy the
input representation encodes consecutive displacements:
\begin{equation}
  \boldsymbol{\delta}_t
    = \mathbf{x}_{t+1} - \mathbf{x}_t,
    \quad t = 1,\ldots,\Tobs{-}1
  \label{eq:delta}
\end{equation}
The prediction target at horizon step~$\tau$ is likewise a displacement:
$\boldsymbol{\delta}_{\Tobs+\tau}
  = \mathbf{x}_{\Tobs+\tau+1} - \mathbf{x}_{\Tobs+\tau}$.
The full input at timestep~$t$ concatenates the displacement with the
contemporaneous velocity:
\begin{equation}
  \mathbf{s}_t
    = \begin{bmatrix} \boldsymbol{\delta}_t \\ \mathbf{v}_t \end{bmatrix}
    \in \mathbb{R}^6
  \label{eq:state}
\end{equation}

Displacements are normalized by a scale constant $\lambda = 2.5$\,m,
chosen so that the 99th percentile of per-step displacement magnitudes
in the training set falls within $[-1,1]^3$.  This encoding provides
three properties: (i)~\emph{translation invariance}, since
$\mathbf{s}_t$ is unchanged by any rigid shift of the trajectory;
(ii)~\emph{bounded input distribution}, since normalization by~$\lambda$
concentrates input mass within a compact region; and
(iii)~\emph{Gaussian compatibility}, since step-to-step displacements of
a smooth trajectory approximate a zero-mean distribution well matched to
the GMM output parameterization.

Future positions are recovered from predicted displacements by cumulative
summation anchored at the last observed position:
\begin{equation}
  \hat{\mathbf{x}}_{\Tobs+\tau}
    = \mathbf{x}_{\Tobs}
      + \sum_{j=1}^{\tau} \hat{\boldsymbol{\delta}}_j \cdot \lambda
  \label{eq:recover}
\end{equation}

\subsection{Multi-Modal Distribution Representation}

At each predicted timestep, the distribution over future displacements
is modeled as a Gaussian Mixture Model (GMM):
\begin{equation}
  p(\mathbf{y}_t \mid \mathcal{X})
    = \sum_{k=1}^{K}
      \pi_k^{(t)}\,
      \mathcal{N}\!\left(
        \mathbf{y}_t \;\middle|\;
        \boldsymbol{\mu}_k^{(t)},\,
        \boldsymbol{\Sigma}_k^{(t)}
      \right)
  \label{eq:gmm}
\end{equation}
where $K$ is the number of mixture components,
$\pi_k^{(t)}$ is the mixture weight for component~$k$ at time~$t$ with
$\sum_k \pi_k^{(t)} = 1$,
$\boldsymbol{\mu}_k^{(t)} \in \mathbb{R}^3$ is the component mean, and
$\boldsymbol{\Sigma}_k^{(t)}
  = \mathrm{diag}(\sigma_{k,x}^{(t)},\,
                   \sigma_{k,y}^{(t)},\,
                   \sigma_{k,z}^{(t)})$
is a diagonal covariance.  The mixture components represent alternative
behavioral modes; their weights encode mode likelihoods, and their
covariances quantify positional uncertainty.  This formulation provides a
downstream planner with a richer signal than a scalar confidence score, where concentrated weights indicate high prediction confidence, and
distributed weights indicate a behavioral decision point at which
conservative planning is appropriate.

\section{METHODOLOGY}
\label{sec:method}

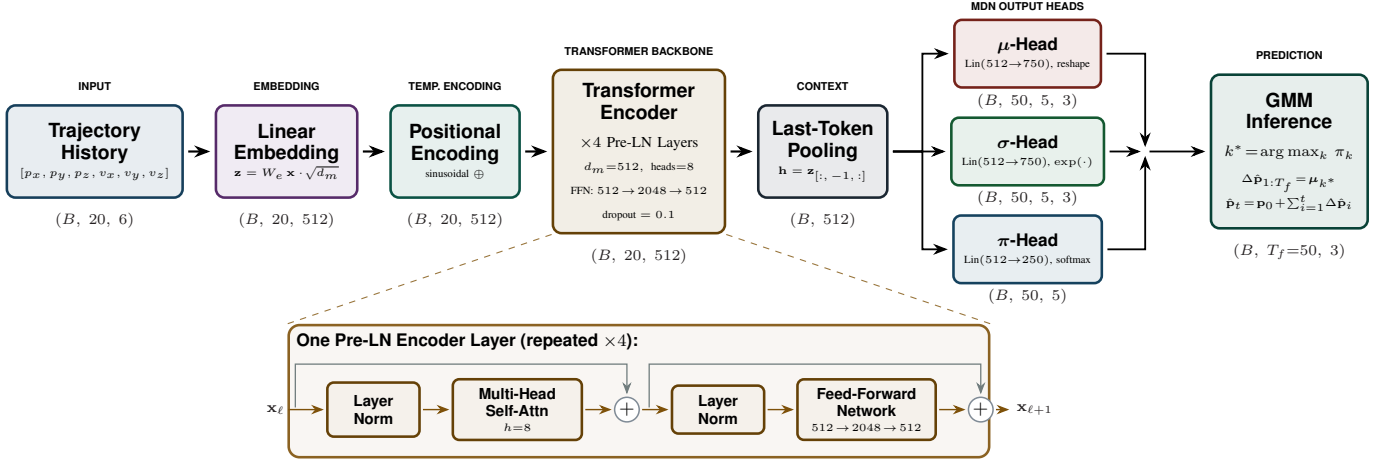
\begin{figure*}[t]
  \centering
  \resizebox{\linewidth}{!}{\begin{tikzpicture}


\node[pipebox={cIn}{2.0cm}, minimum height=1.5cm] (input) at (0,0) {
  Trajectory\\History\\[-2pt]
  \normalfont\tiny $[p_x, p_y, p_z, v_x, v_y, v_z]$
};

\node[pipebox={cEmb}{2.3cm}, minimum height=1.5cm,
      right=0.5cm of input] (embed) {
  Linear\\Embedding\\[-2pt]
  \normalfont\tiny $\mathbf{z} = W_e\,\mathbf{x}\cdot\!\sqrt{d_m}$
};

\node[pipebox={cPE}{2.1cm}, minimum height=1.5cm,
      right=0.5cm of embed] (pe) {
  Positional\\Encoding\\[-2pt]
  \normalfont\tiny sinusoidal $\oplus$
};

\node[pipebox={cEnc}{2.8cm}, minimum height=2.6cm,
      right=0.5cm of pe] (enc) {
  Transformer\\Encoder\\[3pt]
  \normalfont\scriptsize ${\times}4$ Pre-LN Layers\\[1pt]
  \normalfont\tiny $d_m{=}512,\;\text{heads}{=}8$\\[1pt]
  \normalfont\tiny FFN: $512\!\to\!2048\!\to\!512$\\[1pt]
  \normalfont\tiny dropout $= 0.1$
};

\node[pipebox={cPool}{2.1cm}, minimum height=1.5cm,
      right=0.5cm of enc] (pool) {
  Last-Token\\Pooling\\[-2pt]
  \normalfont\tiny $\mathbf{h} = \mathbf{z}_{[:,\,-1,\,:]}$
};


\node[mdnbox={cSig}{2.3cm},
      right=1.0cm of pool] (sig) {
  $\boldsymbol{\sigma}$\textbf{-Head}\\[-1pt]
  \normalfont\tiny Lin$(512{\to}750)$, $\exp(\cdot)$
};

\node[mdnbox={cMu}{2.3cm},
      above=0.45cm of sig] (mu) {
  $\boldsymbol{\mu}$\textbf{-Head}\\[-1pt]
  \normalfont\tiny Lin$(512{\to}750)$, reshape
};

\node[mdnbox={cPi}{2.3cm},
      below=0.45cm of sig] (pi) {
  $\boldsymbol{\pi}$\textbf{-Head}\\[-1pt]
  \normalfont\tiny Lin$(512{\to}250)$, softmax
};


\node[pipebox={cOut}{2.55cm}, minimum height=2.5cm,
      right=1.7cm of sig] (out) {
  GMM\\Inference\\[3pt]
  \normalfont\scriptsize $k^{*}\!=\!\arg\max_{k}\,\pi_{k}$\\[2pt]
  \normalfont\tiny $\Delta\hat{\mathbf{p}}_{1:T_f}\!=\!\boldsymbol{\mu}_{k^{*}}$\\[1pt]
  \normalfont\tiny $\hat{\mathbf{p}}_t\!=\!\mathbf{p}_0\!+\!
                    {\textstyle\sum_{i=1}^{t}}\Delta\hat{\mathbf{p}}_i$
};


\node[shp, below=0.1cm of input] {$(B,\;20,\;6)$};
\node[shp, below=0.1cm of embed] {$(B,\;20,\;512)$};
\node[shp, below=0.1cm of pe]    {$(B,\;20,\;512)$};
\node[shp, below=0.1cm of enc]   {$(B,\;20,\;512)$};
\node[shp, below=0.1cm of pool]  {$(B,\;512)$};
\node[shp, below=-0.05cm of mu]    {$(B,\;50,\;5,\;3)$};
\node[shp, below=-0.05cm of sig]   {$(B,\;50,\;5,\;3)$};
\node[shp, below=-0.05cm of pi]    {$(B,\;50,\;5)$};
\node[shp, below=0.1cm of out]   {$(B,\;T_f{=}50,\;3)$};


\node[stag, above=0.1cm of input] {\textbf{INPUT}};
\node[stag, above=0.1cm of embed] {\textbf{EMBEDDING}};
\node[stag, above=0.1cm of pe]    {\textbf{TEMP.\ ENCODING}};
\node[stag, above=0.1cm of enc]   {\textbf{TRANSFORMER BACKBONE}};
\node[stag, above=0.1cm of pool]  {\textbf{CONTEXT}};
\node[stag] at ($(sig.north)+(0,1.8)$) {\textbf{MDN OUTPUT HEADS}};
\node[stag, above=0.1cm of out]   {\textbf{PREDICTION}};


\draw[far] (input) -- (embed);
\draw[far] (embed) -- (pe);
\draw[far] (pe)    -- (enc);
\draw[far] (enc)   -- (pool);

\draw[bar] (pool.east) -- ++(0.55,0) |- (mu.west);
\draw[bar] (pool.east) --              (sig.west);
\draw[bar] (pool.east) -- ++(0.55,0) |- (pi.west);

\coordinate (mJ) at ($(sig.east)+(0.65,0)$);
\draw[bar] (mu.east)  -| (mJ);
\draw[bar] (sig.east) -- (mJ);
\draw[bar] (pi.east)  -| (mJ);
\draw[bar] (mJ) -- (out.west);


\node[draw=cEnc!90, fill=cEnc!5!white, rounded corners=5pt,
      minimum width=11.4cm, minimum height=2.15cm,
      line width=1.2pt]
      (dbg) at ($(enc.south)+(0,-2.55)$) {};

\node[font=\footnotesize\sffamily\bfseries, color=black,
      anchor=north west, inner sep=4pt]
      at (dbg.north west)
      {One Pre-LN Encoder Layer (repeated ${\times}4$):};

\node[dbox, minimum width=1.5cm, minimum height=0.88cm]
      (d_ln1) at ($(dbg.west)+(1.4, -0.3)$) {Layer\\Norm};

\node[dbox, minimum width=2.1cm, minimum height=0.78cm,
      right=0.5cm of d_ln1] (d_mha) {
  Multi-Head\\Self-Attn\\[-1pt]
  \normalfont\tiny $h{=}8$
};

\node[circ, right=0.5cm of d_mha]  (d_p1) {$+$};

\node[dbox, minimum width=1.5cm, minimum height=0.78cm,
      right=0.5cm of d_p1] (d_ln2) {Layer\\Norm};

\node[dbox, minimum width=2.25cm, minimum height=0.78cm,
      right=0.5cm of d_ln2] (d_ffn) {
  Feed-Forward\\Network\\[-1pt]
  \normalfont\tiny $512\!\to\!2048\!\to\!512$
};

\node[circ, right=0.5cm of d_ffn]  (d_p2) {$+$};

\draw[dar] ($(d_ln1.west)+(-0.65,0)$) -- (d_ln1);
\draw[dar] (d_ln1) -- (d_mha);
\draw[dar] (d_mha) -- (d_p1);
\draw[dar] (d_p1)  -- (d_ln2);
\draw[dar] (d_ln2) -- (d_ffn);
\draw[dar] (d_ffn) -- (d_p2);
\draw[dar] (d_p2)  -- ++(0.5,0);

\draw[res]
  ($(d_ln1.west)+(-0.5,0)$) -- ++(0, 0.74)
  -| (d_p1.north);

\draw[res]
  ($(d_p1.east)+(0.15,0)$)  -- ++(0, 0.74)
  -| (d_p2.north);

\node[shp, left=0.05cm of d_ln1, xshift=-0.5cm]
      {\scriptsize $\mathbf{x}_\ell$};
\node[shp, right=0.25cm of d_p2]
      {\scriptsize $\mathbf{x}_{\ell+1}$};

\draw[dashed, color=cEnc!80, line width=0.5pt]
      (enc.south west) -- (dbg.north west);
\draw[dashed, color=cEnc!80, line width=0.5pt]
      (enc.south east) -- (dbg.north east);

\end{tikzpicture}
}
  \caption{AeroCast architecture. A 6-DoF trajectory history
           $(B, 20, 6)$ is projected to $\dmodel{=}512$ dimensions,
           enriched with sinusoidal positional encoding, and processed
           by $L{=}4$ Pre-LN Transformer encoder layers. The last-token
           context vector feeds three parallel MDN heads that output
           per-waypoint Gaussian mixture parameters
           $(\boldsymbol{\mu}, \boldsymbol{\sigma}, \boldsymbol{\pi})$
           with $K{=}5$ components. At inference, the dominant mixture
           component is selected and cumulative summation yields
           $\Tpred{=}50$ 3-D waypoints at $\Delta t{=}0.1$\,s.}
  \label{fig:architecture}
\end{figure*}

\subsection{Architecture Overview}

The AeroCast architecture comprises three functional blocks: an input
embedding that projects the six-dimensional state $\mathbf{s}_t$ to a
$\dmodel$-dimensional representation, a stack of $L$~Transformer encoder
layers performing temporal self-attention over the observation window,
and three linear output heads that jointly produce the GMM parameters
$\{\boldsymbol{\mu}_k^{(\tau)}\}$,
$\{\boldsymbol{\Sigma}_k^{(\tau)}\}$, and
$\{\pi_k^{(\tau)}\}$ for all $\Tpred \times K$ mixture components in a
single forward pass.  Fig.~\ref{fig:architecture} illustrates the
complete pipeline.  The model uses $\dmodel = 512$, $L = 4$ encoder
layers, $H = 8$ attention heads ($d_k = 64$ per head), feedforward inner
dimension $4\dmodel = 2048$, $K = 5$ GMM components, and dropout
probability $p = 0.1$.  The total parameter count is 13.15\,M.

\subsection{Input Embedding with Positional Encoding}

Each state vector $\mathbf{s}_t \in \mathbb{R}^6$ is projected to a
$\dmodel$-dimensional embedding:
\begin{equation}
  \mathbf{h}_t^{(0)}
    = \sqrt{\dmodel}\;\mathbf{W}_{\mathrm{emb}}\,\mathbf{s}_t
      + \mathbf{e}_t^{\mathrm{pos}}
  \label{eq:embed}
\end{equation}
where $\mathbf{W}_{\mathrm{emb}} \in \mathbb{R}^{\dmodel \times 6}$ is
a learned projection matrix and
$\mathbf{e}_t^{\mathrm{pos}}$ is the sinusoidal positional encoding~\cite{vaswani2017attention}:
\begin{align}
  \mathbf{e}_t^{\mathrm{pos}}[2i]
    &= \sin\!\left(\frac{t}{10000^{2i/\dmodel}}\right)
    \label{eq:pe_sin} \\
  \mathbf{e}_t^{\mathrm{pos}}[2i{+}1]
    &= \cos\!\left(\frac{t}{10000^{2i/\dmodel}}\right)
    \label{eq:pe_cos}
\end{align}
for $i = 0,\ldots,\dmodel/2{-}1$.  The $\sqrt{\dmodel}$ scaling prevents
the fixed positional signal from dominating the learned content embedding
at large model widths.  Because self-attention is
permutation-equivariant, the positional encoding is the sole mechanism by
which the model distinguishes temporal order within the observation
window~\cite{vaswani2017attention}.  The ablation in Section~\ref{sec:ablation} quantifies the
impact of removing it.

\subsection{Transformer Encoder}

The embedded sequence
$\{\mathbf{h}_t^{(0)}\}_{t=1}^{\Tobs}$
is processed through $L$ identical encoder layers.  Each layer applies
pre-layer normalization (Pre-LN)~\cite{xiong2020layer}:
\begin{align}
  \tilde{\mathbf{h}}_t^{(\ell)}
    &= \mathbf{h}_t^{(\ell-1)}
       + \mathrm{MHA}\!\bigl(
           \mathrm{LN}(\mathbf{H}^{(\ell-1)})
         \bigr)_t
    \label{eq:preln_attn} \\[3pt]
  \mathbf{h}_t^{(\ell)}
    &= \tilde{\mathbf{h}}_t^{(\ell)}
       + \mathrm{FFN}\!\bigl(
           \mathrm{LN}(\tilde{\mathbf{h}}_t^{(\ell)})
         \bigr)
    \label{eq:preln_ffn}
\end{align}
where $\mathbf{H}^{(\ell-1)} \in \mathbb{R}^{\Tobs \times \dmodel}$ is
the sequence matrix and $\mathrm{LN}(\cdot)$ denotes layer
normalization.  The multi-head attention operation uses $H$ parallel
heads of dimension $d_k = \dmodel / H$:
\begin{equation}
  \mathrm{MHA}(\mathbf{H})
    = \mathrm{Concat}(\mathrm{head}_1,\ldots,\mathrm{head}_H)\,
      \mathbf{W}^O
  \label{eq:mha}
\end{equation}
\begin{equation}
  \mathrm{head}_h
    = \mathrm{softmax}\!\left(
        \frac{\mathbf{Q}_h \mathbf{K}_h^\top}{\sqrt{d_k}}
      \right) \mathbf{V}_h
  \label{eq:sdpa}
\end{equation}
with
$\mathbf{Q}_h{=}\mathbf{H}\mathbf{W}_h^Q$,
$\mathbf{K}_h{=}\mathbf{H}\mathbf{W}_h^K$,
$\mathbf{V}_h{=}\mathbf{H}\mathbf{W}_h^V$,
where
$\mathbf{W}_h^Q, \mathbf{W}_h^K, \mathbf{W}_h^V
  \in \mathbb{R}^{\dmodel \times d_k}$
and $\mathbf{W}^O \in \mathbb{R}^{\dmodel \times \dmodel}$.
The scaled dot-product produces a $\Tobs \times \Tobs$ affinity matrix,
giving every timestep $O(1)$ access to every other timestep in the
window, compared with $O(\Tobs)$ for recurrent encoders.

The position-wise feed-forward network applies a two-layer MLP with ReLU
activation:
\begin{equation}
  \mathrm{FFN}(\mathbf{h})
    = \mathbf{W}_2\,\mathrm{ReLU}(\mathbf{W}_1\mathbf{h}+\mathbf{b}_1)
      + \mathbf{b}_2
  \label{eq:ffn}
\end{equation}
with $\mathbf{W}_1 \in \mathbb{R}^{4\dmodel \times \dmodel}$ and
$\mathbf{W}_2 \in \mathbb{R}^{\dmodel \times 4\dmodel}$.  Dropout is
applied after both the MHA and FFN outputs, before each residual
addition.

After the $L$-th layer, a terminal layer normalization is applied to the
output at the last temporal position to produce the context vector:
\begin{equation}
  \mathbf{c}
    = \mathrm{LN}\!\left(\mathbf{h}_{\Tobs}^{(L)}\right)
    \in \mathbb{R}^{\dmodel}
  \label{eq:context}
\end{equation}
This vector serves as the sole input to the three output heads; no
recurrent or autoregressive decoding is employed.

\subsection{GMM Output Heads}

Three linear projections map $\mathbf{c}$ to the GMM parameters for all
timesteps and components simultaneously:
\begin{align}
  \hat{\boldsymbol{\mu}}
    &= \mathbf{W}_\mu\,\mathbf{c}
    \;\in\; \mathbb{R}^{\Tpred \times K \times 3}
    \label{eq:mu_head} \\
  \log\hat{\boldsymbol{\sigma}}
    &= \mathbf{W}_\sigma\,\mathbf{c}
    \;\in\; \mathbb{R}^{\Tpred \times K \times 3}
    \label{eq:sig_head} \\
  \hat{\mathbf{z}}_\pi
    &= \mathbf{W}_\pi\,\mathbf{c}
    \;\in\; \mathbb{R}^{\Tpred \times K}
    \label{eq:pi_head}
\end{align}
Standard deviations and mixture weights are obtained via:
\begin{align}
  \hat{\sigma}_{k,d}^{(\tau)}
    &= \max\!\bigl(
         e^{\log\hat{\sigma}_{k,d}^{(\tau)}},\;
         \sigma_{\min}
       \bigr)
    \label{eq:sigma_act} \\[3pt]
  \hat{\pi}_k^{(\tau)}
    &= \frac{
         e^{\hat{z}_{\pi,k}^{(\tau)}}
       }{
         \sum_{j=1}^{K} e^{\hat{z}_{\pi,j}^{(\tau)}}
       }
    \label{eq:pi_act}
\end{align}
The variance floor $\sigma_{\min} = 0.05$\,m matches the Vicon
measurement noise level.  This bound prevents degenerate point-mass
predictions and ensures that predicted confidence intervals remain
physically interpretable.  The full output dimensionality per forward
pass is $\Tpred \times K \times (3{+}3{+}1)$ scalars; with $\Tpred=50$
and $K=5$ this yields $50 \times 5 \times 7 = 1{,}750$ outputs.

\subsection{Training Objective}
\label{sec:training}

The model is trained to maximize the log-likelihood of observed
displacements under the predicted GMM.  The batch-averaged negative
log-likelihood is:
\begin{equation}
\small
  \mathcal{L}_{\mathrm{NLL}}
    = -\frac{1}{N\Tpred}
      \sum_{n=1}^{N} \sum_{\tau=1}^{\Tpred}
      \log \sum_{k=1}^{K}
        \hat{\pi}_k^{(\tau)}\,
        \mathcal{N}\!\bigl(
          \boldsymbol{\delta}_\tau^{(n)}
          \;\big|\;
          \hat{\boldsymbol{\mu}}_k^{(\tau)},\,
          \hat{\boldsymbol{\Sigma}}_k^{(\tau)}
        \bigr)
  \label{eq:nll}
\end{equation}
computed via the standard log-sum-exp identity for numerical stability.

Pure NLL training with $K > 1$ admits a degenerate solution in which all
components spread to cover the data support without any single component
being geometrically close to the ground truth.  This degeneracy is
consequential because ADE and FDE at evaluation time are computed from
the dominant-weight component
$k^* = \arg\max_k \hat{\pi}_k^{(\tau)}$.  To break it without
collapsing the predicted variance (as a large MSE weight would), a small
auxiliary mean-squared error term is applied to~$k^*$:
\begin{equation}
  \mathcal{L}_{\mathrm{MSE}}
    = \frac{1}{N\Tpred}
      \sum_{n=1}^{N} \sum_{\tau=1}^{\Tpred}
      \bigl\|
        \hat{\boldsymbol{\mu}}_{k^*}^{(\tau)}
        - \boldsymbol{\delta}_\tau^{(n)}
      \bigr\|^2
  \label{eq:mse}
\end{equation}
The combined objective is:
\begin{equation}
  \mathcal{L}(\boldsymbol{\theta})
    = \mathcal{L}_{\mathrm{NLL}}
      + \lambda_{\mathrm{MSE}}\,\mathcal{L}_{\mathrm{MSE}}
  \label{eq:loss}
\end{equation}
with $\lambda_{\mathrm{MSE}} = 0.15$, chosen so that the MSE term
contributes roughly 10--15\% of the total loss magnitude.  NLL remains
the dominant signal, preserving the proper-scoring-rule property that
encourages well-spread predictive distributions.

\subsubsection{Curriculum Sequence-Length Scheduling}

Sliding-window sequences drawn from short trajectory segments can
exhibit irregular leading-edge behavior that destabilizes early
training.  A curriculum schedule addresses this by gradually increasing
the effective observation length during the first
$E_{\mathrm{warm}} = 20$ epochs:
\begin{equation}
  T_{\mathrm{eff}}(e)
    = \min\!\Bigl(\Tobs,\;
        \max\!\bigl(5,\;
          \lfloor \Tobs \cdot e / E_{\mathrm{warm}} \rfloor
        \bigr)
      \Bigr)
  \label{eq:curriculum}
\end{equation}
At epoch~$e$, the model processes only the most recent
$T_{\mathrm{eff}}(e)$ timesteps of each window; the full 20-step window
is used for all subsequent epochs.  Additive Gaussian noise
($\sigma_{\mathrm{aug}} = 0.02$\,m) and random scale perturbation
(uniform in $[0.95, 1.05]$) are applied to inputs during the warmup
phase.

\section{DATASET}
\label{sec:dataset}

We constructed the dataset combining real-world
flights captured in a motion-capture arena with synthetic trajectories
generated by an extension of our prior synthetic data framework
SynTraG~\cite{syed2026syntrag}. This section describes the physical
collection methods, the synthetic augmentation procedure, the
preprocessing pipeline, and the resulting dataset statistics.

\subsection{Physical Data Collection}

\subsubsection{Hardware Platform}
 
Real trajectories were collected with a Crazyflie~2.1 nano-quadrotor
(Bitcraze~AB, $92 \times 92 \times 29$~mm, 27~g with a 250~mAh single-cell
LiPo battery installed) flown as the obstacle agent in a Vicon-equipped
indoor arena of dimensions $6 \times 4 \times 2$~m. The arena is
instrumented with twelve infrared cameras providing rigid-body 6-DoF pose
at 100~Hz with sub-millimeter spatial accuracy. The platform was chosen for its small inertia and high agility, which
enable the abrupt heading reversals and multi-axis maneuvers required to
exercise the multi-modal regime of the prediction problem.

\subsubsection{Trajectory Categories}
 
Trajectories were collected across nine motion categories chosen to span
the kinematic envelope an autonomous vehicle is likely to encounter when
sharing airspace with non-cooperative agents. The categories fall into
three groups: \textit{periodic} (Circle, Oval, Figure-8), which approximate
loitering and patrolling behaviors with strong temporal regularity;
\textit{non-trivially 3-D} (Helix, Trefoil, Lissajous, Staircase), which
couple horizontal and vertical motion in ways that challenge planar
prediction; and \textit{aperiodic} (Star, Random), which contain abrupt
heading reversals and unstructured motion that resist physics-based
filter modeling. Reference paths for each category were generated as
parametric curves and tracked under closed-loop position control;
parameter values (radius, amplitude, frequency, altitude offset) were
randomized across flights to span the dataset's diversity in speed and
extent.

\subsection{Synthetic Data Augmentation}

Approximately 43\% of the data corpus is synthetic, generated
by an extension of the SynTraG framework~\cite{syed2026syntrag} that
expands the four kinematic primitives described in the original work to
the nine categories used here. Synthetic trajectories follow the same
parametric structure as the physical reference paths and are generated
with randomized speed profiles, radii, altitudes, and noise levels to
fill regions of the kinematic state space that are under-represented in
the physical collection, particularly in the high-speed and high-altitude
regimes that the limited arena volume cannot accommodate. Heteroscedastic
Gaussian noise consistent with the Vicon noise floor
($\sigma_\text{ms} = 0.05$~m, the same value used as the sigma floor in
the prediction head) is added to synthetic samples to match the
distribution of measurement noise in the real data. Synthetic and real
samples are pooled before splitting and treated identically downstream;
no synthetic-versus-real label is provided to the model.

\subsection{Preprocessing and Sliding-Window Sampling}
\label{sec:preproc}
 
Each recorded trajectory is downsampled from 100\,Hz to 10\,Hz to match
the prediction-system update rate. Velocities are reconstructed from the
position stream by the centered finite difference with Gaussian smoothing
defined in Eq.~\eqref{eq:velocity}. Sliding windows are extracted with stride one:
each window comprises a 20-step observation followed by a 50-step
prediction target, corresponding to 2~s of history and 5~s of horizon
at 10~Hz. Stride one is chosen because temporally adjacent windows differ
in their prediction targets by a single timestep at the leading edge,
which exposes the model to maximum target diversity per trajectory; the
implied correlation between adjacent samples is bounded by the trajectory-level
split described next. The state vector at each input timestep is the
displacement-velocity pair $\mathbf{s}_t = [\bd_t^\top, \bv_t^\top]^\top
\in \mathbb{R}^6$ defined in Section~\ref{subsec:disp_enc}.
 
The train/validation partition is performed at the trajectory level:
each recording is assigned in its entirety to one split, ensuring that
no observation from a validation trajectory appears in training and that
each window's 5-second prediction target is statistically independent of
all training samples.  The split ratio is 85/15 with a fixed random
seed.  Stratification by trajectory category ensures that all nine
motion types appear in both splits in proportion to their frequency in
the full corpus.
 
\subsection{Dataset Statistics}
\label{sec:stats}
 
Table~\ref{tab:dataset} summarizes the full corpus by category, reporting
the sliding-window sample count, mean per-window RMS speed, peak speed,
and the percentage contributed by physical and synthetic flights. The
total sample count is $90\,116$ from
113 unique trajectory recordings. Sample counts are roughly comparable across the
nine categories; the balanced distribution ensures that all motion types
receive adequate representation during training and that aggregate
metrics are not dominated by any single category. Mean per-window
speeds range from $0.5$~m/s on Staircase
trajectories (slow vertical maneuvers) to $0.79$~m/s
on the more agile periodic categories.

\begin{table}[t]
  \centering
  \caption{Dataset summary by trajectory category. Speed is the mean
           per-sequence RMS speed over the 20-step (2\,s) observation
           window. Real/Synth indicates the percentage of sequences
           contributed by physical flights versus synthetic generation.}
  \label{tab:dataset}
  \setlength{\tabcolsep}{4pt}
  \begin{tabular}{@{}lrrrrr@{}}
    \toprule
    Category & Sequences & Mean speed & Max speed & Real & Synth. \\
             &           & (m/s)      & (m/s)     & (\%) & (\%)  \\
    \midrule
    Circle       & 10,248 & 0.78 & 1.43 & 64 & 36 \\
    Oval         & 10,062 & 0.79 & 1.37 & 62 & 38 \\
    Figure-8     &  9,874 & 0.74 & 1.77 & 60 & 40 \\
    Helix        & 10,136 & 0.79 & 1.57 & 57 & 43 \\
    Trefoil      & 10,410 & 0.71 & 1.90 & 58 & 42 \\
    Lissajous    &  9,748 & 0.73 & 1.34 & 50 & 50 \\
    Staircase    &  9,512 & 0.50 & 0.88 & 46 & 54 \\
    Star         & 10,286 & 0.65 & 1.36 & 62 & 38 \\
    Random       &  9,840 & 0.55 & 1.04 & 53 & 47 \\
    \midrule
    Total        & 90,116 & 0.69 & 1.41 & 57 & 43 \\
    \bottomrule
  \end{tabular}
\end{table}
\section{EXPERIMENTAL SETUP}
\label{sec:setup}

\subsection{Baseline Methods}

Four baseline architectures are trained under conditions identical to
AeroCast.  All baselines share the same MDN output head ($K{=}5$
mixture components), the same training objective
as described in Section~\ref{sec:training}, and the same data splits.

\textbf{GRU-MDN}~\cite{nacar2025vector} uses a three-layer
unidirectional GRU encoder with hidden dimension~512.  This baseline
isolates the effect of replacing sequential temporal encoding with
global self-attention.

\textbf{LSTM-MDN}~\cite{lstm_mdn} uses a three-layer LSTM encoder with
hidden dimension~512.  The explicit cell state provides a distinct
inductive bias relative to GRUs for retaining information over the full
20-step observation window.

\textbf{BiGRU-MDN}~\cite{zhi2021bigru} uses a three-layer bidirectional
GRU with hidden dimension~512.  The backward pass gives the encoder
access to future-looking context within the observation window, at the
cost of roughly doubling the recurrent parameter count relative to
GRU-MDN.

\textbf{MLP-MDN}~\cite{Guo_2023_mlp_mdn} uses a three-layer MLP with
hidden dimensions $(2048, 2048, 1024)$ that receives the flattened
20-step observation window (120~input features) and maps it directly to
GMM parameters.  This baseline tests whether a feed-forward architecture
with sufficient capacity can implicitly capture temporal structure.

Parameter counts are 4.75\,M (GRU-MDN), 6.07\,M (LSTM-MDN), 13.17\,M
(BiGRU-MDN), and 7.63\,M (MLP-MDN).  AeroCast uses 13.15\,M parameters,
closely matching BiGRU-MDN.

\subsection{Evaluation Metrics}

Six metrics are used to evaluate trajectory accuracy, distributional
quality, and calibration.

\textit{Average Displacement Error (ADE)} is the mean Euclidean distance
between predicted and ground-truth positions over all future timesteps:
\begin{equation}
  \mathrm{ADE}
    = \frac{1}{N\Tpred}
      \sum_{n=1}^{N}\sum_{\tau=1}^{\Tpred}
      \left\|
        \hat{\mathbf{x}}_{\Tobs+\tau}^{(n)}
        - \mathbf{x}_{\Tobs+\tau}^{(n)}
      \right\|_2
  \label{eq:ade}
\end{equation}

\textit{Final Displacement Error (FDE)} is the Euclidean distance at the
last predicted timestep:
\begin{equation}
  \mathrm{FDE}
    = \frac{1}{N}
      \sum_{n=1}^{N}
      \left\|
        \hat{\mathbf{x}}_{\Tobs+\Tpred}^{(n)}
        - \mathbf{x}_{\Tobs+\Tpred}^{(n)}
      \right\|_2
  \label{eq:fde}
\end{equation}

\textit{minADE@$K$ and minFDE@$K$} are best-of-$K$ variants obtained by
drawing $K{=}5$ trajectory samples from the predicted GMM and reporting
the minimum ADE or FDE across samples.  These metrics reward coverage of
the ground truth by any mixture component.

\textit{Negative Log-Likelihood (NLL)} is the mean per-timestep negative
log-likelihood under the predicted GMM, as defined in
Eq.~\eqref{eq:nll}.  Lower values indicate higher probability assigned
to observed future displacements.

\textit{Continuous Ranked Probability Score (CRPS)} is an
energy-score-based proper scoring rule that jointly rewards distributional
accuracy and calibration~\cite{gneiting2007proper}:
\begin{equation}
\small
  \mathrm{CRPS}
    = \frac{1}{NK\Tpred}
      \sum_{n,\tau}\!\left(
        \mathbb{E}_{\hat{Y}}
          \|\hat{Y} - \mathbf{y}_\tau^{(n)}\|
        - \tfrac{1}{2}\,
          \mathbb{E}_{\hat{Y},\hat{Y}'}
          \|\hat{Y} - \hat{Y}'\|
      \right)
  \label{eq:crps}
\end{equation}

estimated via Monte Carlo samples from the predicted GMM.

\textit{Expected Calibration Error (ECE)} measures agreement between
predicted uncertainty and empirical coverage.  For $z$-score thresholds
$z_\ell$ spanning $[0.1, 3.0]$, the nominal coverage is
$p_\ell = 2\Phi(z_\ell) - 1$ (where $\Phi$ is the standard normal CDF)
and the empirical coverage $\hat{p}_\ell$ is the fraction of
ground-truth displacements falling within $z_\ell$ predicted standard
deviations of the predicted mean:
\begin{equation}
  \mathrm{ECE}
    = \frac{1}{|\mathcal{Z}|}
      \sum_{\ell} |p_\ell - \hat{p}_\ell|
  \label{eq:ece}
\end{equation}
A perfectly calibrated model achieves $\mathrm{ECE} = 0$.  All six
metrics are computed on the held-out validation split; lower is better
for each.

\subsection{Training Protocol}

All five models are trained with the combined NLL and mode-anchoring
objective defined in Eq.~\eqref{eq:loss}
($\lambda_{\mathrm{MSE}} = 0.15$).  Optimization uses AdamW~\cite{loshchilov2018decoupled} with
$\beta_1 = 0.9$, $\beta_2 = 0.999$, and weight decay $10^{-5}$.  The batch size is 128 for all five models. The
initial learning rate is $\eta_0 = 10^{-4}$ with a minimum of
$\eta_{\min} = 10^{-6}$.  Gradients are clipped at $\ell_2$ norm~0.5.
Training runs for up to 300~epochs with early stopping at patience
120~epochs based on validation NLL.

The learning rate schedule is architecture-dependent.  AeroCast uses
CosineAnnealingLR~\cite{loshchilov2017sgdr} with $T_{\max} = 300$.  The four baselines use
CosineAnnealingWarmRestarts with $T_0 = 20$ and
$T_{\mathrm{mult}} = 2$; the periodic warm restarts empirically improve
convergence for the shallower recurrent architectures.

\subsection{Implementation Details}

All projection matrices (embedding, attention, feed-forward, and output
heads) are initialized with Xavier uniform
initialization~\cite{glorot2010understanding}.  The sinusoidal positional
encoding is fixed (non-learnable); layer normalization modules use
standard learnable affine parameters.  The PyTorch implementation uses
\texttt{TransformerEncoderLayer} with Pre-LN and a terminal
\texttt{LayerNorm} after all $L$~layers, as described in
Eq.~\eqref{eq:context}.  All experiments are conducted on a single NVIDIA
GPU.  The total parameter count for AeroCast is 13.15\,M, of which
12.61\,M reside in the four Transformer encoder layers and the
remainder is split across the input embedding and three output heads.

 \begin{table}[t]
  \centering
  \caption{Quantitative comparison on 13,500 held-out validation
           sequences (5-second, 50-step horizon). All models use
           $K{=}5$ mixture components and matched training recipes.
           mADE/mFDE@5 are best-of-5 sampled trajectories.
           All metrics lower is better. \textbf{Bold}: best per column.}
  \label{tab:comparison}
  \setlength{\tabcolsep}{2.2pt}
  \renewcommand{\arraystretch}{1.05}
  \footnotesize
  \begin{tabular}{@{}l
                   S[table-format=1.3]
                   S[table-format=1.3]
                   S[table-format=1.3]
                   S[table-format=1.3]
                   S[table-format=-2.2]
                   S[table-format=1.3]
                   S[table-format=1.3]@{}}
    \toprule
    {Method} &
    {ADE} & {FDE} & {mADE} & {mFDE} &
    {NLL} & {CRPS} & {ECE} \\
    & {[m]} & {[m]} & {@5 [m]} & {@5 [m]} &
    & {[m]} & \\
    \midrule
    
    GRU-MDN         & 0.127 & 0.211 & 0.122 & 0.185 & -10.31 & 0.169          & 0.674 \\
    LSTM-MDN        & 0.149 & 0.238 & 0.146 & 0.220 &  -9.37 & 0.197          & 0.676 \\
    BiGRU-MDN       & \bfseries 0.111 & \bfseries 0.190 & \bfseries 0.108 & \bfseries 0.175 & -10.06 & 0.145 & 0.669 \\
    MLP-MDN         & 0.125 & 0.219 & 0.115 & 0.183 & \bfseries -11.40 & 0.157 & \bfseries \textbf{0.645} \\
    \midrule
    \textbf{AeroCast (ours)} & \textbf{0.055} & \textbf{0.099} & \textbf{0.052} & \textbf{0.088} & \textbf{-13.28} & \textbf{0.049} & 0.647  \\
    \bottomrule
  \end{tabular}
\end{table}
\section{RESULTS AND ANALYSIS}
\label{sec:results}

\subsection{Quantitative Comparison}
 
Table~\ref{tab:comparison} reports all evaluation metrics on the held-out
validation split across nine trajectory types, evaluated over a 5-second
(50-step) prediction horizon.  All five models use $K{=}5$ mixture
components and are trained under the protocol described in
Section~\ref{sec:setup}.
 
AeroCast achieves an ADE of 0.055\,m and an FDE of 0.099\,m, reducing
both metrics by approximately 50\% relative to BiGRU-MDN
(ADE\,=\,0.111\,m, FDE\,=\,0.190\,m), which is the strongest baseline among the baselines.  The best-of-5 metrics follow the same
ordering: mADE@5 of 0.052\,m and mFDE@5 of 0.088\,m for AeroCast,
compared with 0.108\,m and 0.175\,m for BiGRU-MDN.  The small gap
between single-mode and best-of-$K$ results (0.055 vs.\ 0.052\,m ADE)
indicates that the dominant mixture component already tracks the ground
truth closely; the remaining components contribute supplementary coverage
rather than compensating for a misaligned primary mode.
 
AeroCast also records the lowest NLL ($-13.28$) and CRPS (0.049\,m).
The NLL margin over the next-best method, MLP-MDN ($-11.40$), is
1.88~nats per timestep, and the remaining baselines fall further behind.
On CRPS, AeroCast attains roughly one-third the value of BiGRU-MDN
(0.049 vs.\ 0.145\,m), which is consistent with predicted distributions
that are both accurate in location and appropriately spread.
 
Calibration, measured by ECE, is more tightly clustered across methods.
MLP-MDN records 0.645, marginally below AeroCast at 0.647 while other baselines fall within a 0.03
band.  This narrow spread suggests that the shared training elements,
particularly the sigma floor at $\sigma_{\min}{=}0.05$\,m and the
NLL-dominated loss, govern calibration more than the choice of
encoder.  The ECE values across all methods (0.64--0.68) leave room for
improvement through post-hoc recalibration or temperature scaling.
 
Among the baselines themselves, BiGRU-MDN achieves the lowest
trajectory error (ADE\,=\,0.111\,m), which is consistent with its bidirectional
access to the observation window.  LSTM-MDN places last on both ADE and
FDE despite a larger parameter budget than GRU-MDN (6.07M vs.\ 4.75M),
which suggests that the explicit cell-state mechanism does not
confer an advantage at the 20-step observation length used here.  MLP-MDN, which
flattens the full window into a single vector, has achieved competitive NLL
($-11.40$) and the best ECE (0.645), though its trajectory-error
performance (ADE\,=\,0.125\,m) lags behind the recurrent models with
temporal inductive biases.

\begin{figure}[t]
    \centering
    \includegraphics[width=\linewidth]{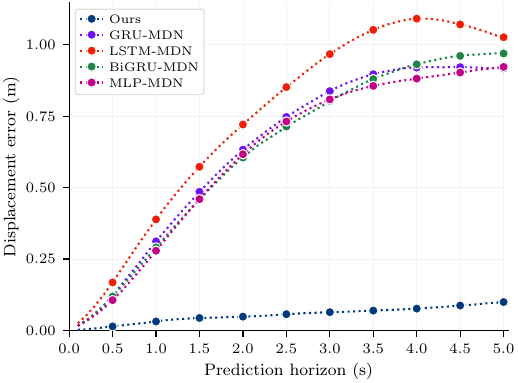}
    \caption{AeroCast position error at increasing prediction horizons
           over the full validation split.}
    \label{fig:horizon}
\end{figure}
 
\subsection{Prediction Horizon Analysis}
 
Fig.~\ref{fig:horizon} plots mean displacement error as a function of
prediction horizon for all five methods.  Error increases monotonically
with horizon length for every model, as expected from displacement-based
cumulative prediction.  AeroCast maintains the lowest error at each
evaluated horizon, and the gap relative to the baselines widens beyond
the 2\,s mark.
 
At short horizons ($<$1\,s), all methods produce low errors that are
within a narrow range of each other: near-future positions are tightly
constrained by the current kinematic state regardless of encoder
architecture.  Beyond 2\,s, the recurrent baselines diverge from the
ground truth at a higher rate than AeroCast.  This pattern is consistent
with the hypothesis that global self-attention propagates early maneuver
signatures across the full observation window more directly than the
sequential hidden-state chain in recurrent encoders, though confirming
this interpretation would require attention-map analysis beyond the
scope of this evaluation.
 
\begin{figure}
    \centering
    \includegraphics[width=\linewidth]{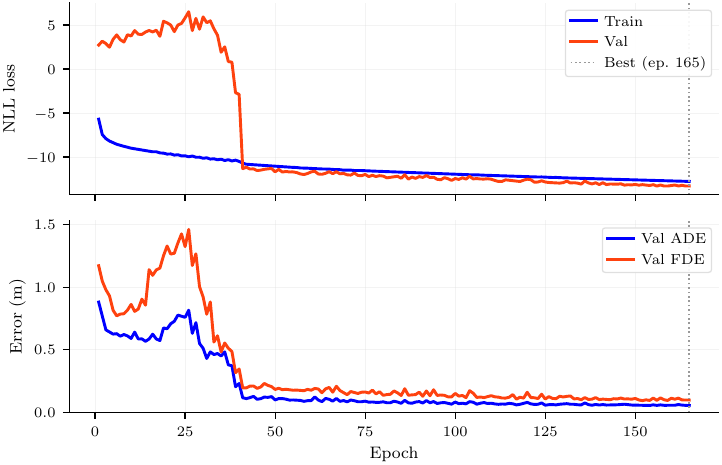}
    \caption{Validation and Training Loss Training Loss convergence over the epochs.}
    \label{fig:convergence}
\end{figure}

Fig.~\ref{fig:convergence} shows the training and validation loss
curves.  The validation NLL decreases steadily and stabilizes without
the oscillatory divergence sometimes observed in pure-NLL training of
mixture models~\cite{bishop1994mdn}.  The combined objective and sigma floor appear to
regularize the optimization effectively.

\begin{table*}[t]
  \centering
  \caption{Per-trajectory-type ADE / FDE (m) on the held-out validation split
           (5-second, 50-step horizon). \textbf{Bold} = best per type per metric.}
  \label{tab:per_type}
  \setlength{\tabcolsep}{4pt}
  \begin{tabular}{@{}lcccccccccc@{}}
    \toprule
    & \multicolumn{2}{c}{\textbf{AeroCast}} &
      \multicolumn{2}{c}{GRU-MDN} &
      \multicolumn{2}{c}{LSTM-MDN} &
      \multicolumn{2}{c}{BiGRU-MDN} &
      \multicolumn{2}{c}{MLP-MDN} \\
    \cmidrule(lr){2-3}\cmidrule(lr){4-5}\cmidrule(lr){6-7}\cmidrule(lr){8-9}\cmidrule(lr){10-11}
    Type & ADE & FDE & ADE & FDE & ADE & FDE & ADE & FDE & ADE & FDE \\
    \midrule
    Trefoil   & \textbf{0.060} & \textbf{0.121} & 0.141 & 0.261 & 0.165 & 0.292 & 0.122 & 0.233 & 0.138 & 0.270 \\
    Helix     & \textbf{0.044} & \textbf{0.085} & 0.103 & 0.183 & 0.120 & 0.206 & 0.089 & 0.165 & 0.092 & 0.204 \\
    Star      & \textbf{0.069} & \textbf{0.198} & 0.161 & 0.423 & 0.189 & 0.478 & 0.140 & 0.381 & 0.157 & 0.441 \\
    Oval      & \textbf{0.048} & \textbf{0.087} & 0.113 & 0.189 & 0.132 & 0.234 & 0.097 & 0.170 & 0.110 & 0.196 \\
    Figure-8  & \textbf{0.061} & \textbf{0.113} & 0.142 & 0.243 & 0.167 & 0.273 & 0.124 & 0.219 & 0.140 & 0.252 \\
    Circle    & \textbf{0.040} & \textbf{0.092} & 0.095 & 0.197 & 0.109 & 0.222 & 0.081 & 0.177 & 0.093 & 0.206 \\
    Random    & \textbf{0.109} & \textbf{0.386} & 0.254 & 0.824 & 0.297 & 0.931 & 0.221 & 0.540 & 0.249 & 0.857 \\
    Lissajous & \textbf{0.069} & \textbf{0.116} & 0.161 & 0.249 & 0.189 & 0.281 & 0.140 & 0.223 & 0.158 & 0.259 \\
    Staircase & \textbf{0.059} & \textbf{0.133} & 0.138 & 0.285 & 0.162 & 0.321 & 0.120 & 0.256 & 0.136 & 0.296 \\
    \bottomrule
  \end{tabular}
\end{table*}
 
\subsection{Per-Trajectory-Type Analysis}
\label{sec:per_type}
 
Table~\ref{tab:per_type} disaggregates ADE and FDE by trajectory
category for all five methods.  AeroCast achieves the lowest ADE and
FDE on every category.  The margin over the strongest baseline varies
with the kinematic characteristics of each type.
 
On periodic categories (Circle, Oval, Figure-8, Trefoil), AeroCast
records ADE values between 0.040\,m and 0.061\,m.
BiGRU-MDN, the strongest baseline on these types, ranges from 0.081\,m to 0.124\,m.  The difference
shows that the Transformer encoder captures periodic phase structure
at least as effectively as the bidirectional recurrent model, without
relying on an explicit recurrent inductive bias.
 
Categories with non-trivial 3-D coupling (Helix, Lissajous, Staircase)
show a similar pattern.  AeroCast records 0.044\,m on Helix, 0.069\,m
on Lissajous, and 0.059\,m on Staircase; the corresponding BiGRU-MDN
values are 0.089\,m, 0.140\,m, and 0.120\,m.  Staircase trajectories,
characterized by slow vertical maneuvers with discrete altitude
transitions, present a qualitatively different displacement distribution
from the smooth periodic types, yet the relative improvement remains
comparable.
 
The largest absolute errors for all methods occur on Random trajectories.
AeroCast obtains ADE of 0.109\,m and FDE of 0.386\,m; BiGRU-MDN
reaches 0.221\,m and 0.540\,m.  The FDE reduction on Random (28.5\%) is
smaller than the reductions exceeding 45\% observed on periodic types,
which reflects the greater intrinsic difficulty of long-horizon prediction
on aperiodic motion with genuine behavioral multi-modality.  Star
trajectories, which contain abrupt heading reversals at each arm tip,
produce the second-highest AeroCast FDE (0.198\,m), though this still
represents a 48\% reduction relative to BiGRU-MDN (0.381\,m).

\begin{figure*}
    \centering
    \includegraphics[width=\textwidth]{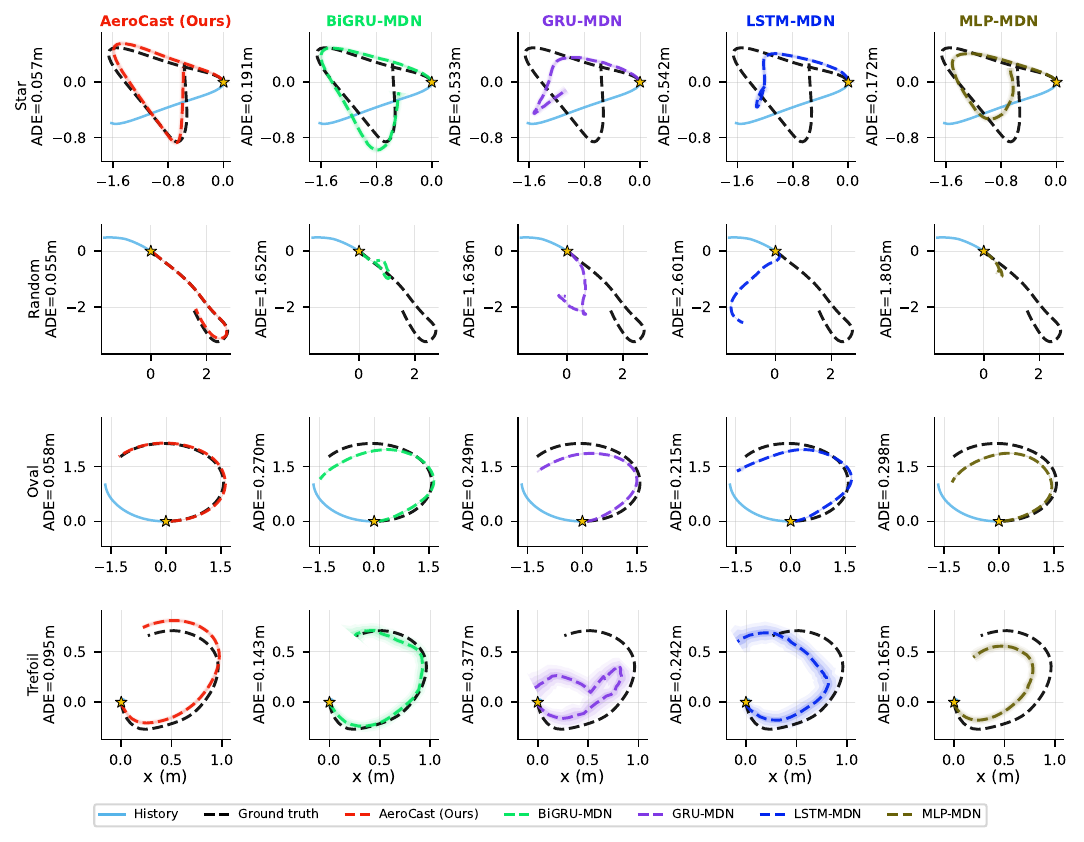}
    \caption{Qualitative comparison of AeroCast with the baseline models on various trajectory categories.}
    \label{fig:qualitative_comp}
\end{figure*}
 
\begin{figure}
    \centering
    \includegraphics[width=\linewidth]{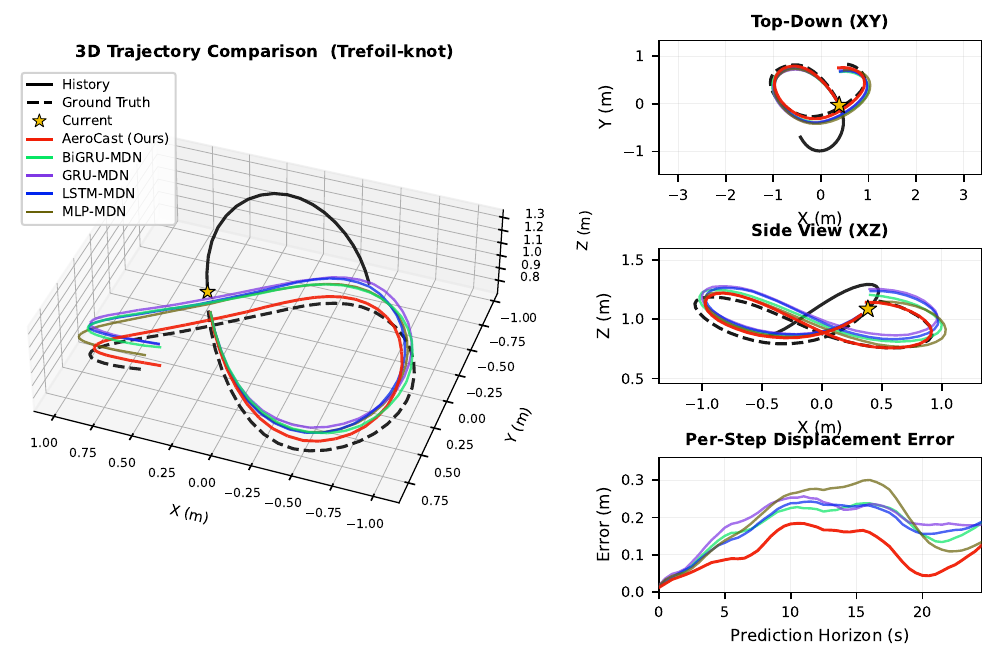}
    \caption{Qualitative comparison of AeroCast with the baseline models on a Trefoil trajectory.}
    \label{fig:trefoil_comp}
\end{figure}

Fig.~\ref{fig:qualitative_comp} corroborates these trends visually
across representative samples from multiple categories.  On structured
trajectories such as Trefoil and Oval, AeroCast predictions closely
follow the ground-truth geometry, and the baselines exhibit visible
drift at curvature maxima.  On Star trajectories, all methods incur
larger errors at the arm tips, but AeroCast recovers the correct heading
more rapidly after each reversal.  Fig.~\ref{fig:trefoil_comp}
presents a detailed view of a single Trefoil sequence with 3-D overlay,
planar projections, and per-step displacement error.  The per-step error
curve shows that AeroCast maintains lower error throughout the
prediction window.

 
\subsection{Ablation Study}
\label{sec:ablation}
 
Table~\ref{tab:ablation} reports results from a systematic ablation
of four design dimensions: mixture cardinality, model capacity,
positional encoding, and input feature set.  Each variant modifies
exactly one dimension relative to the full model ($K{=}5$, $d{=}512$,
$L{=}4$, sinusoidal positional encoding, 6-D displacement-velocity
input).

\begin{table}[t]
  \centering
  \caption{%
    Ablation study on the held-out validation split.
    \textbf{Bold} = best in column.
    $K$: mixture components; $d$: model width; $L$: transformer layers;
    PE: positional encoding; Inp: input features (6=pos+vel, 3=pos only).%
  }
  \label{tab:ablation}
  \setlength{\tabcolsep}{3pt}
  \scriptsize
  \renewcommand{\arraystretch}{1.2}
  \begin{tabular}{l|ccccc|c|c|c|c}
    \toprule
    & \multicolumn{5}{c|}{Architecture}
    & ADE$\downarrow$
    & FDE$\downarrow$
    & NLL$\downarrow$
    & ECE$\downarrow$ \\[1mm]
    Variant & $K$ & $d$ & $L$ & PE & Inp
    & [m] & [m] & & \\
    \midrule
    \textbf{Full model (ours)} & 5 & 512 & 4 & \checkmark & 6 & \textbf{0.055} & \textbf{0.099} & -13.285 & 0.647 \\
    \midrule
    \multicolumn{10}{l}{\textit{Number of mixture components $K$}} \\
    $K=3$             & 3  & 512 & 4 & \checkmark & 6 & 0.076 & 0.130 & -12.806          & \textbf{0.6405} \\
    $K=10$            & 10 & 512 & 4 & \checkmark & 6 & 0.101          & 0.163          & \textbf{-14.021}          & 0.6665 \\
    $K=1$ (uni.)      & 1  & 512 & 4 & \checkmark & 6 & 0.133          & 0.213          & -12.377          & 0.6536 \\
    \midrule
    \multicolumn{10}{l}{\textit{Model depth \& width}} \\
    Tiny ($d$=128, $L$=2)   & 5 & 128 & 2 & \checkmark & 6 & 0.247 & 0.513 & -12.354          & 0.6662 \\
    Medium ($d$=256, $L$=4) & 5 & 256 & 4 & \checkmark & 6 & 0.079 & 0.179 & -12.898          & 0.6569 \\
    Small ($d$=256, $L$=2)  & 5 & 256 & 2 & \checkmark & 6 & 0.112 & 0.223 & -12.634          & 0.6628 \\
    \midrule
    \multicolumn{10}{l}{\textit{Positional encoding}} \\
    No pos. enc.      & 5 & 512 & 4 & \texttimes & 6 & 0.118 & 0.410 & -13.146 & 0.6600 \\
    \midrule
    \multicolumn{10}{l}{\textit{Input features}} \\
    Pos.-only (3-D)   & 5 & 512 & 4 & \checkmark & 3 & 0.098 & 0.187 & -11.684          & 0.7883 \\
    \bottomrule
  \end{tabular}
\end{table}

\subsubsection{Mixture Cardinality}
 
Reducing $K$ from 5 to 1 (unimodal Gaussian) raises ADE from 0.055\,m
to 0.133\,m and worsens NLL from $-13.285$ to $-12.377$ which shows
that the multi-modal output head provides more capacity for
covering the kinematic diversity present in the dataset.  At $K{=}3$,
ADE increases to 0.076\,m, but ECE improves to 0.6405, the lowest
among all ablation configurations.  A fewer-component mixture thus
yields slightly better-calibrated uncertainty at the expense of reduced
trajectory accuracy.  At $K{=}10$, NLL improves to $-14.021$, but ADE rises to 0.101\,m and ECE degrades to 0.6665.  The
additional components improve the global density fit but do not
individually anchor to the ground-truth path, so the argmax-weight
mode used for ADE/FDE evaluation is less precise.  $K{=}5$ offers the
best trade-off between distributional fit and dominant-mode accuracy.
 
\subsubsection{Model Capacity}
 
The Tiny variant ($d{=}128$, $L{=}2$, 1.85M parameters) incurs ADE of
0.247\,m and FDE of 0.513\,m, a fourfold degradation relative to the
full model.  Increasing to $d{=}256$, $L{=}2$ (Small) reduces ADE to
0.112\,m; adding depth ($d{=}256$, $L{=}4$, Medium) brings it to
0.079\,m.  The full model ($d{=}512$, $L{=}4$) achieves 0.055\,m, a
further 30\% reduction over the Medium variant.  NLL improvements are
more modest across this progression ($-12.354$ for Tiny to $-13.285$
for Full) that suggest that additional capacity improves the geometric
precision of the dominant mixture component more than the overall
distributional fit.
 
\subsubsection{Positional Encoding}
 
Removing the sinusoidal positional encoding increases ADE from 0.055\,m
to 0.118\,m and FDE from 0.099\,m to 0.410\,m.  The asymmetry is
notable as ADE roughly doubles, but FDE increases by a factor of four.
Without explicit temporal ordering, the model retains some ability to
produce reasonable displacement predictions on average by relying on
token content alone, but it cannot maintain trajectory coherence through
the full 50-step horizon, as reflected in the FDE
degradation.  NLL changes only modestly ($-13.285$ to $-13.146$),
so the contribution of positional encoding is primarily to geometric
accuracy rather than distributional fit.
 
\subsubsection{Input Features}
 
Removing velocity and retaining only 3-D displacements raises ADE from
0.055\,m to 0.098\,m (78\% increase) and FDE from 0.099\,m to
0.187\,m.  NLL drops from $-13.285$ to $-11.684$, and ECE rises from
0.647 to 0.7883. The velocity signal conveys instantaneous heading and speed information
that cannot be reliably recovered from a 20-step displacement history,
particularly on trajectories where speed changes precede directional
changes.  The ECE degradation is the most severe observed in any
single-factor ablation, which implies that velocity context is critical for both positional accuracy and for producing
uncertainty estimates whose spread reflects actual kinematic variability.

\subsection{Inference Latency}
 
Table~\ref{tab:complexity} reports model size and per-sample inference
latency measured at batch size 1 on a single NVIDIA GPU.
AeroCast completes a forward pass in 0.100\,ms, within the 10\,ms
budget for a 100\,Hz prediction loop and comparable to BiGRU-MDN
(0.125\,ms).  GRU-MDN and LSTM-MDN are faster (0.042 and 0.056\,ms)
due to smaller parameter counts.  MLP-MDN is the fastest at 0.004\,ms.
All five methods fall well within a single control cycle at 100\,Hz;
the computational bottleneck in an onboard deployment would reside in
the upstream perception pipeline rather than in the prediction step.

 \begin{table}[t]
  \centering
  \caption{Comparison of model complexity and inference latency (with batch size 1 over 50 runs after warm-up)}
  \label{tab:complexity}
  \setlength{\tabcolsep}{4pt}
  \begin{tabular}{@{}lrrr@{}}
    \toprule
    Method & Parameters & Size (MB) & Latency (ms) \\
    \midrule
    AeroCast (ours) & 13.15M & 54.0 & 0.100 \\
    GRU-MDN    & \textbf{4.75M}  &  \textbf{19.0} & 0.042 \\
    LSTM-MDN   & 6.07M  &  24.3 & 0.056 \\
    BiGRU-MDN  & 13.17M &  52.7 & 0.125 \\
    MLP-MDN    & 7.63M  &  30.5 & \textbf{0.004} \\
    \bottomrule
  \end{tabular}
\end{table}

\subsection{Discussion}
 
Across all seven evaluation metrics and all nine trajectory categories,
AeroCast either matches or improves upon every baseline.  The trajectory-error
and distributional advantages are consistent regardless of whether the
motion is periodic, 3-D-coupled, or aperiodic.
 
On periodic trajectories, where a recurrent model could in principle
memorize the repeating displacement pattern, AeroCast still attains
lower ADE.  It is due to the fact that self-attention over the full
20-step window allows the encoder to identify the instantaneous phase
of the motion more directly than a recurrent encoder that must propagate
phase information through intermediate hidden states.  On aperiodic
types, the same global connectivity may help the model detect early
kinematic signatures of heading changes that sequential processing would
attenuate.  Both of these analyses are our own hypotheses that would require
attention-weight analysis to confirm.
 
The ablation results rank velocity input and model capacity as the two
largest contributors to prediction quality.  Positional encoding plays a
significant role and its removal has a limited effect on ADE
but quadruples FDE, that means temporal ordering is essential
specifically for long-horizon trajectory coherence.  The finding that
velocity removal degrades ECE from 0.647 to 0.7883, more severely
than any other single-factor ablation, points to a concrete mechanism:
without instantaneous speed information, the model cannot differentiate
slow-moving obstacles (whose future positions occupy a compact region)
from fast-moving ones (whose future positions are more dispersed),
resulting in miscalibrated predictive spreads.
 
Several limitations qualify the conclusions drawn above.  All
experiments use a single dataset comprising indoor quadrotor flights and
parametrically generated synthetic trajectories; generalization to
outdoor environments, larger vehicles, or biological aerial agents
remains untested. The evaluation considers
single-obstacle scenarios; multi-obstacle settings, where inter-agent
interactions introduce additional behavioral modes, would require
architectural extensions.

\section{CONCLUSION}
\label{sec:conclusion}

This paper presented AeroCast, a probabilistic trajectory prediction
framework for non-cooperative aerial obstacles that pairs a Pre-LN
Transformer encoder with a Mixture Density Network output head.  Three
design choices underpin the approach: a consecutive displacement encoding
that enforces translation invariance and bounded input statistics, a
combined NLL and mode-anchoring training objective with a physically
motivated sigma floor, and curriculum sequence-length scheduling for
early training stability.

AeroCast is evaluated against four baselines on a hybrid real-and-synthetic
quadrotor dataset spanning nine motion categories. AeroCast reduces ADE
and FDE by approximately 50\% relative to the strongest baseline
(BiGRU-MDN) and achieves the lowest NLL and CRPS among all compared
methods.  These improvements hold across all nine trajectory types,
with the largest absolute gains on aperiodic and structurally complex
categories.  Ablation analysis identifies velocity input as the single
most important feature for both trajectory accuracy and calibration
quality, and positional encoding as essential for maintaining prediction
coherence over the full 5-second horizon.  AeroCast inference latency
(0.100\,ms per sample) remains within the real-time budget for onboard
deployment at 100\,Hz.

Future work includes extension to multi-obstacle settings
with inter-agent interaction modeling, online adaptation to specific
obstacle agents observed over time, and integration with a
receding-horizon planner that consumes the predicted GMM distributions
as probabilistic obstacle constraints.

\section*{ACKNOWLEDGMENT}

Research was sponsored by the Army Research Office and was accomplished under Grant Number W911NF-23-1-0186. The views and conclusions contained in this document are those of the authors and should not be interpreted as representing the official policies, either expressed or implied, of the Army Research Office or the U.S. Government. The U.S. Government is authorized to reproduce and distribute reprints for Government purposes notwithstanding any copyright notation herein. It was also supported by by the Geospatial Computer Science Program at Texas A\&M University-Corpus Christi.

\bibliographystyle{IEEEtran}
\bibliography{IEEEexample}

\begin{IEEEbiography}[{\includegraphics[width=1in,height=1.25in,clip,keepaspectratio]{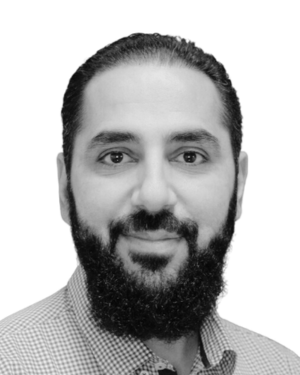}}]{Syed Izzat Ullah} is a Ph.D. candidate in Computer Science at Texas A\&M University-Corpus Christi, Texas, USA. His research focuses on risk-aware multi-robot motion planning, integrating reinforcement learning, transfer learning, and optimization to improve autonomous navigation in dynamic and uncertain environments. Before starting his Ph.D. studies, he earned a B.S. in Telecommunication Engineering from Balochistan University of IT, Engineering \& Management Sciences, Pakistan, in $2016$, followed by an M.S. in Electrical Engineering (Robotics \& Control Systems) from Lahore University of Management Sciences (LUMS), Pakistan, in $2019$. His research has been presented at leading robotics conferences and has held research positions at the National Center of Robotics \& Automation, Pakistan, as well as a visiting researcher role at the Robotics Research Lab, TU Kaiserslautern, Germany.
\end{IEEEbiography}%

\begin{IEEEbiography}[{\includegraphics[width=1in,height=1.25in,clip,keepaspectratio]{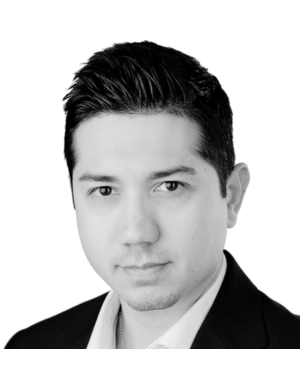}}]{Jos\'e Baca} is an Associate Professor in the Department of Engineering at Texas A\&M University-Corpus Christi, USA. He earned his B.S. in Electrical Engineering from Instituto Tecnologico de Matamoros, Mexico, his M.Sc. in Mechatronics from University of Applied Sciences in Aachen, Germany, his Ph.D. in Automation and Robotics, from Universidad Politecnica de Madrid, Spain, and worked as a Postdoctoral Researcher in the Computer Science Department at University of Nebraska at Omaha, USA. His research interests include the development of coordination and control strategies for Unmanned Autonomous Systems and the integration of Modular Systems across different domains such as in robotics, search and rescue, coastal resilience, space, industry, agriculture, healthcare, and education. He has organized and co-chaired international conferences and workshops, and has been involved in projects funded by agencies such as DoD, NSF, USDA, and NASA.
\end{IEEEbiography}

\end{document}